\documentclass[letterpaper, 10 pt, conference]{ieeeconf}  

\IEEEoverridecommandlockouts                              

\usepackage{graphics} 
\usepackage[hyperindex=true,colorlinks, citecolor=blue,menucolor=blue]{hyperref} 
\hypersetup{
	colorlinks=true, 
	citecolor=blue,
	filecolor=blue,
	linkcolor=blue,
	urlcolor=blue,
	linktoc=all,     
}
\usepackage{epsfig} 
\usepackage{mathptmx} 
\usepackage{times} 
\usepackage{amsmath} 
\usepackage{amssymb}  
\usepackage{multirow} 
\usepackage{array} 
\usepackage{setspace} 
\usepackage{colortbl} 
\usepackage{tabulary}
\usepackage{etoolbox}
\usepackage{hhline}
\usepackage{hyperref} 

\usepackage[usenames,dvipsnames]{xcolor}
\makeatletter
\let\MYcaption\@makecaption
\makeatother

\usepackage[font=footnotesize]{subcaption}

\makeatletter
\let\@makecaption\MYcaption
\makeatother

\usepackage{float}

\makeatletter
\def\mathcolor#1#{\@mathcolor{#1}}
\def\@mathcolor#1#2#3{%
	\protect\leavevmode
	\begingroup
	\color#1{#2}#3%
	\endgroup
}
\makeatother

\let\orgautoref\autoref
\renewcommand{\autoref}{%
	\def\sectionautorefname{Section}%
	\def\figureautorefname{Fig.}%
	\orgautoref%
}

\title{\LARGE \bf
	Incremental learning of high-level concepts by imitation 
}

\author{Mina Alibeigi$^{1}$, Majid Nili Ahmadabadi$^{2}$ and Babak Nadjar Araabi$^{2}$
\thanks{$^{1}$Mina Alibeigi is with Cognitive Systems Lab., School of Electrical and Computer Engineering, College of Engineering, University of Tehran, Tehran, Iran 
	{\tt\small minaalibeigi@gmail.com}; {\tt\small m.alibeigi@ut.ac.ir}}%
\thanks{$^{2}$Majid Nili Ahmadabadi and Babak Nadjar Araabi are with Cognitive Systems Lab., Control and Intelligent Processing Center of Excellence, School of Electrical and Computer Engineering, College of Engineering, University of Tehran, Tehran, Iran. They are also with School of Cognitive Sciences, Institute for Research in Fundamental Sciences (IPM), Tehran, Iran
        {\tt\small \{mnili, araabi\}@ut.ac.ir}}%
\thanks{$^{*}$The elaborate version of this paper is available at \cite{alibeigi2017fast}.}
}

\begin{document}

\maketitle
\thispagestyle{empty}
\pagestyle{empty}

\begin{abstract}
Nowadays, robots become a companion in everyday life. To be well-accepted by humans, robots should efficiently understand meanings of their partners' motions and body language, and respond accordingly. Learning concepts by imitation brings them this ability in a user-friendly way.

This paper presents a fast and robust model for Incremental Learning of Concepts by Imitation (ILoCI). In ILoCI, observed multimodal spatio-temporal demonstrations are incrementally abstracted and generalized based on both their perceptual and functional similarities during the imitation. In this method, perceptually similar demonstrations are abstracted by a dynamic model of mirror neuron system. An incremental method is proposed to learn their functional similarities through a limited number of interactions with the teacher. Learning all concepts together by the proposed memory rehearsal enables robot to utilize the common structural relations among concepts which not only expedites the learning process especially at the initial stages, but also improves the generalization ability and the robustness against discrepancies between observed demonstrations.

Performance of ILoCI is assessed using standard LASA handwriting benchmark data set. The results show efficiency of ILoCI in concept acquisition, recognition and generation in addition to its robustness against variability in demonstrations.
\end{abstract}

\begin{keywords} 
	Concepts, imitation learning, humanoid robots, social human-robot interaction 
\end{keywords}

\section{INTRODUCTION}
Nowadays, along with the advances in sensing and learning techniques, applications of robots have been extended from controlled to unstructured and complex environments \cite{kemp2007challenges, billard2008handbook}. Company of robots in humans' daily life have caused lots of difficulties in designing and programming them, since they should operate in complex environments with unpredictable or time-varying dynamics and interact with humans \cite{kemp2007challenges, billard2008handbook, lopes2010abstraction}. Moreover, ordinary users generally do not have enough expertise to program robots for new tasks \cite{kemp2007challenges, billard2008handbook}. In addition, to gain acceptance as an intelligent companion in our everyday life, robots should be sociable. They should understand the meanings of their partners' motions and body language, and respond accordingly. These requirements and limitations specify the necessity of developing socially interactive learning methods for robots to enable them to effectively cope with new environments and tasks instead of being manually pre-programmed \cite{kemp2007challenges, billard2008handbook, lopes2010abstraction}.

Inspiring by the efficient social learning methods in animals and humans (e.g. mimicry, emulation and goal emulation), researchers proposed natural and user-friendly ways to teach robots, which is called robot programming by demonstration or imitation learning \cite{kemp2007challenges, billard2008handbook, lopes2010abstraction}. Although all the social learning methods from the high-level knowledge transfer to the low-level exact regeneration of observed demonstrations are mistakenly known as imitation, but there are stark differences between them \cite{billard2008handbook, lopes2010abstraction, call2002three}. In the high-level methods, in contrast to the low-level ones, understanding the teacher's intentions along with regenerating actions are required \cite{billard2008handbook, lopes2010abstraction, call2002three}. In this level, also called "true imitation", skills are abstracted in a generalized symbolic representation. Abstraction, conceptualization and symbolization are bases of true imitation. They bring decreased state-space as one of the requirements of real applications in addition to expediting the knowledge transfer from one agent or situation to another \cite{billard2008handbook, hajimirsadeghi2013conceptual, call2002three, inamura2004embodied, kadone2006segmentation}. 

In recent years, abstraction and symbolization have received a great deal of attention by researchers in the field of imitation learning \cite{hajimirsadeghi2013conceptual, inamura2004embodied, kadone2006segmentation, tani2004self, ito2004line, kulic2008incremental}. A considerable portion of the proposed methods inspired by the presumed role of mirror neurons in imitative behaviors of animals and humans \cite{hajimirsadeghi2013conceptual, inamura2004embodied, tani2004self, ito2004line}. Tani et al. \cite{tani2004self, ito2004line, tani2005interacting} proposed an offline bio-inspired method called recurrent neural networks with parametric biases (RNNPB), as a model of mirror neuron system. In this model, the observed spatio-temporal demonstrations are learned and abstracted by the network based on their perceptual properties. Moreover, Inamura et al. \cite{inamura2004embodied} proposed another bio-inspired imitation learning method inspiring the mirror neurons and mimesis theory \cite{donald1991origins}. In this model, hidden markov models (HMMs) are used for abstracting and symbolizing the observed human motions as well as for recognizing and generating them. Demonstrations of different motion patterns are manually grouped and encoded into distinct HMMs in an offline manner. The number of HMMs representing different behaviors should be known a priori; which is not suitable for real applications. Moreover, the method is not incremental, meaning that it does not give robot the ability to learn concepts gradually and autonomously in cooperation with the partners in order to keep itself socially competent.

Considering these shortcomings into account, some methods were proposed for incremental learning of human motions \cite{kadone2006segmentation, kulic2008incremental}. One of the prominent representative algorithms is proposed by Kadone and Nakamura \cite{kadone2006segmentation}. This model affords autonomous segmentation, abstraction, memorization and recognition of demonstrated motions using associative neural networks. Kulic et al. \cite{kulic2008incremental} proposed another well-known incremental and autonomous imitation learning method for acquisition, symbolization, recognition and hierarchical organization of whole body motion patterns using Factorial HMMs. 

Although, in all the mentioned studies \cite{inamura2004embodied, kadone2006segmentation, tani2004self, ito2004line, kulic2008incremental, tani2005interacting}, only the perceptual similarity among observed demonstrations are addressed for abstraction and symbolization, but there are some perceptually different concepts that have the same functional effects or semantic meanings, called relational concepts \cite{hajimirsadeghi2013conceptual, mahmoodian2013hierarchical, zentall2002categorization, mobahi2007biologically}. These concepts cannot be specified merely based on their perceptual properties and an extra information is needed to acquire them \cite{hajimirsadeghi2013conceptual, mahmoodian2013hierarchical, zentall2002categorization, mobahi2007biologically}. They are highly prevalence in humans' social interactions and their everyday life; for instance, disparate gestures to convey the meaning of "Hello" in different cultures. Therefore, functional categorization of observed demonstrations is also indispensable for robots coexisting with humans. However, despite the prevalence of relational concepts, not enough researches carried out in this field.

To the best of our knowledge, only a limited number of researches has been proposed for learning and abstracting concepts based on both their perceptual and functional properties \cite{hajimirsadeghi2013conceptual, mahmoodian2013hierarchical, mobahi2007biologically, hajimirsadeghi2011conceptual}. One of the basic models is proposed by Mobahi et al. \cite{mobahi2007biologically}. The model is just applicable for learning concepts from single observations, and is not directly extendible to continuous sequences of observations. In contrast, the proposed methods by Hajmirsadeghi et al. \cite{hajimirsadeghi2013conceptual, hajimirsadeghi2011conceptual} are applicable 
for learning concepts from spatio-temporal motion sequences using both perceptual and functional properties. In these models, each relational concept is represented by a group of distinct HMM prototypes that each symbolize a different perceptual variant of that concept. Separated modeling of prototypes in these models \cite{hajimirsadeghi2013conceptual, hajimirsadeghi2011conceptual}, leads to neglecting their common structural relations and consequently each prototype should relearn the common knowledge again. Therefore, the learning speed decreases and more observations are needed for generalization. This is in contradiction to the main idea of the imitation learning that supports expediting the autonomous training of robots using the minimum number of demonstrations. 

Considering the mentioned requirements and limitations, this paper presents a gradual and incremental learning algorithm to abstract and generalize the observed multimodal spatio-temporal demonstrations based on both their perceptual and functional characteristics during the imitation. The proposed method comprises low-level and high-level modules. The low-level module abstracts the observed spatio-temporal demonstrations based on their perceptual properties using an RNNPB network \cite{tani2004self, ito2004line, tani2005interacting}. The high-level module acquires relational concepts based on the formed perceptual prototypes and the perceived teacher's feedbacks. The proposed memory rehearsal procedure enables the robot to gradually extract and utilize the common structural relations among concepts. Therefore, the learning process is expedited especially at the initial stages and the generalization capability is improved as well as the robustness against noise and variations among observed demonstrations.
\section{ILoCI: The proposed method for Incremental Learning of Concepts by Imitation}
\begin{figure}[tpb]
	\includegraphics[width=\linewidth]{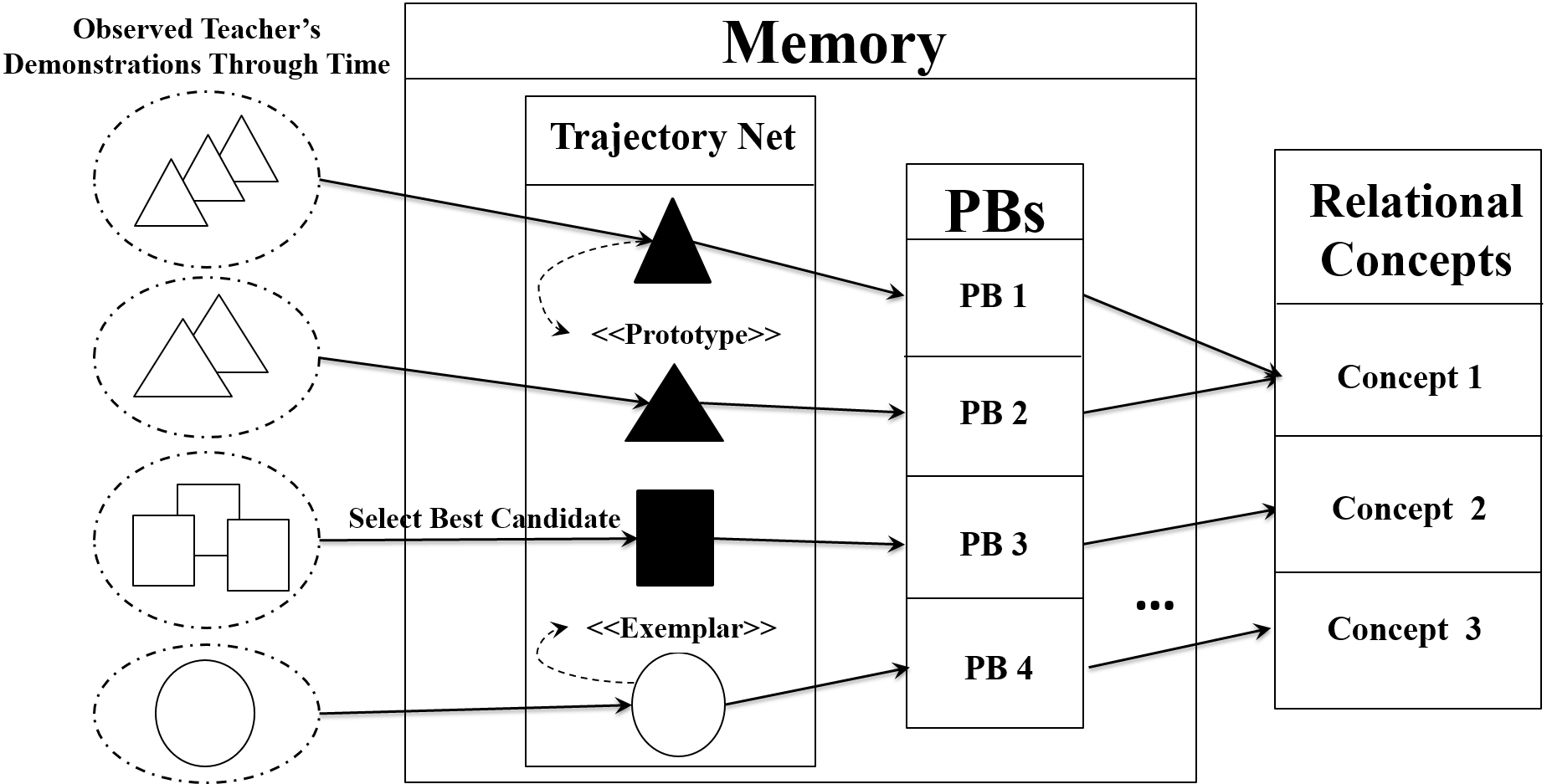}
	\caption{Mem and consolidated exemplars and prototypes. Filled shapes show prototypes and unfilled shapes depict demonstrations and exemplars.}
	\label{Mem_Figure}
\end{figure}

In a nutshell, ILoCI has a low-level and a high-level module. The low-level module of ILoCI is a dynamic model of mirror neuron systems, called RNNPB, which abstracts the observed multimodal spatio-temporal demonstrations as perceptual concepts. For more details on RNNPB refer to \cite{tani2004self, ito2004line, tani2005interacting}. It automatically assigns a PB vector to each acquired perceptual prototype. The acquired PB vectors can be exemplars or prototypes based on their associated information in the high-level module. An exemplar PB vector stands for only one demonstration and a prototype PB vector is the medoid of demonstrations with sufficient perceptual similarity. All the exemplar and prototype PB vectors along with their associated information are stored in a memory in the high-level module, called "\textit{Mem}" (see \autoref{Mem_table}). A relational concept is defined as a set of perceptually variant exemplars and prototypes in the memory that have same functional properties. The high-level module learns the relational concepts by employing the low-level module and the acquired teacher's feedbacks through interactions. \autoref{Mem_Figure} illustrates the relations among exemplars, prototypes and concepts. In the sequel, ILoCI is explained in more details. 
\begin{table}[tbp]
	\caption{Definition of Some Symbols}
	\label{Mem_table}
	\centering
	\scriptsize
	\bgroup
	\def\arraystretch{1.1}	
	\resizebox{\linewidth}{!}{%
		\begin{tabular}{|>{\centering\arraybackslash}m{0.05\linewidth}|>{\centering\arraybackslash}m{0.18\linewidth}|>{\centering\arraybackslash}m{0.09\linewidth}|m{0.57\linewidth}|}
			\hline
			Name                 & Constituents           & Type             & Description                                                                                                          \\ \hline
			\multirow{9}{*}{Mem} & \textit{nPrototypes}   & \textit{Int}     & Number of all consolidated exemplars and prototypes in \textit{Mem}.                                                          \\ \cline{2-4} 
			& \textit{TrajectoryNet} & \textit{Network} & The RNNPB that abstracts and symbolizes the consolidated exemplars and prototypes.                                   \\ \cline{2-4} 
			& \textit{PBs}           & \textit{Set}     & PB vectors assigned to the learned demonstrations by \textit{TrajectoryNet.}                                                  \\ \cline{2-4} 
			& \textit{PBs\_rec}      & \textit{Set}     & PB vectors generated by \textit{TrajectoryNet} when recognizing each learned demonstration.                                   \\ \cline{2-4} 
			& \textit{numSamples}    & \textit{Set}     & Number of sufficiently similar observed demonstrations associated to each consolidated exemplar or prototype in \textit{Mem}. \\ \cline{2-4} 
			& \textit{numSteps}      & \textit{Set}     & Number of time steps of each consolidated demonstrations in \textit{Mem}.                                                     \\ \cline{2-4} 
			& \textit{initialInfo}   & \textit{Set}     & Initial configuration of each consolidated demonstration in \textit{Mem}.                                                     \\ \cline{2-4} 
			& \textit{conceptLabels} & \textit{Set}     & Concept label assigned to each of the consolidated demonstrations in \textit{Mem}.                                            \\ \cline{2-4} 
			& \textit{generationError}        & \textit{Set}     & Error of regenerating each consolidated demonstration in \textit{Mem}.                                                        \\ \hline
		\end{tabular}
	}
}
\end{table}
\subsection{Learning Phase}
The main procedure of ILoCI is an iterative cycle triggered by the advent of a new teacher's demonstration. After perceiving a new demonstration, the smoothing, scaling and fitting post-processes are activated consecutively. Then, the processed demonstration is fed into the inverse kinematics function to compute its corresponding motor data. Afterwards, the obtained sensory and motor data are input into the low-level module to recognize the corresponding concept.

After recognizing the concept, the robot performs an action in response to the teacher and receives a reinforcement signal accordingly. Receiving a reward, the robot uses the observed demonstration to update or develop its memory. In contrast, in the case of punishment, robot tries other available concepts until receiving a reward. If none of the former concepts in the robot's memory are proper for the new demonstration, a new concept will be generated and consolidated in memory. In this way, the robot gradually and incrementally learns and develops the relational concepts in imitation of the teacher to increase its lifetime rewards. In following, steps are described in more details.
\subsubsection{Perceiving new demonstration}
At first an observation from the teacher goes through pre-processing. Details are described in \autoref{results_discussion}. After preparing the observed motion sequence, the robot tries to find its associated concept. To do so, the observed motion sequence in terms of sensory and motor data, is fed into \textit{Mem.TrajectoryNet} and the value of $PB_{obs}$ is computed for it by back propagating and minimizing the error between the target and the predicted values of sensory and motor data. Afterwards, in order to find the most similar consolidated $PBs\_rec$ in memory, the computed value of $PB_{obs}$ is compared with the untried associated $PBs\_rec$ values of the consolidated concepts in memory.  

The concept of the most similar consolidated exemplar or prototype is selected as the guessed concept of the novel observed demonstration ($CL_{obs}$) and is added to the set of the currently tried concepts ($Q_{tried}$). Then, in response to the teacher, the robot executes the action with the lowest generation error among the actions with $CL_{obs}$ concept in its memory. After performing the selected action, robot receives a feedback (reward or punishment) from the teacher, which helps it to adjust its concepts. According to the received reinforcement signal, robot faces three situations:

\textit{Receiving positive reinforcement signal with high similarity between the compared PB vectors}: A positive feedback shows that the robot has found the concept of the newly observed demonstration correctly. Moreover, it is an evidence of a highly similar exemplar or prototype for the that demonstration in the robot's memory and fulfills the need of relearning. Therefore, the most similar consolidated demonstration in the robot's memory is strengthened as a potential candidate for the new observed demonstration. 

\textit{Receiving positive reinforcement signal with low similarity between the compared PB vectors}: In this case, $CL_{obs}$ has been found correctly but there is no enough perceptually similar exemplar or prototype for that demonstration in the memory. Therefore, the robot should learn a new prototype for that relational concept in its memory and consolidate it through memory rehearsal. After a while, memory may be overpopulated with perceptually similar exemplars and prototypes. Therefore, these demonstrations should be abstracted and clustered in order to select the best representatives of their counterpart clusters. Thus, a complete link hierarchical agglomerative clustering is called when a new exemplar of a concept is added to the memory while the number of samples of both prototypes and exemplars of that concept exceeds $Num_{threshold}$. Afterwards, final valid clusters are selected based on two criteria. First, the number of demonstrations should exceed a predefined threshold with at least one exemplar in the cluster. Second, the mean of the pairwise Euclidean distances among PB vectors within the cluster should be less than $D_{cutoff}$ (\ref{equ:clustering_threshold}). This threshold is computed based on the mean ($\mu$) and the standard deviation ($\sigma$) of the pairwise Euclidean distances across all vectors in the clusters of the desired concept. 
\begin{equation}
D_{cutoff} = \mu - K_{cutoff}\,*\,\sigma
\label{equ:clustering_threshold}
\end{equation}

\begin{figure}[tpb]	
	\includegraphics[width=\linewidth]{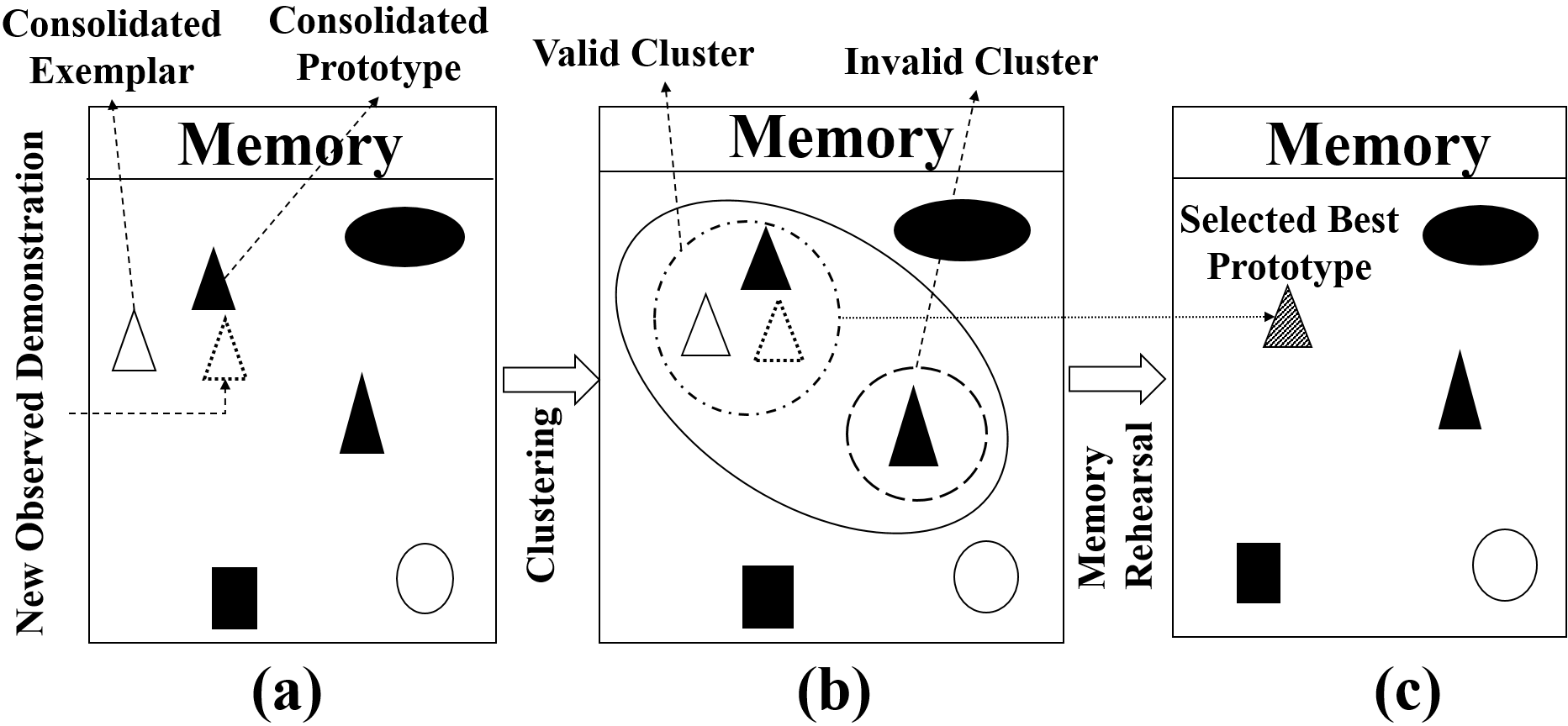}
	\caption{An illustration for the clustering process of the triangle concept, (a) new demonstration is observed and consolidated in the memory as a new perceptual representation of the triangle concept, (b) clustering is performed and valid clusters are determined based on the validity criteria, (c) the medoid of the exemplars and prototypes in the valid cluster is substituted for other members through memory rehearsal.}
	\label{Clustering_fig}
\end{figure}

In (\ref{equ:clustering_threshold}), $K_{cutoff}$ is a predefined parameter that controls the granularity level of the algorithm. Higher values of $K_{cutoff}$ lead to more number of specific prototypes; while lower values bring more general prototypes. However, all variant perceptual prototypes of a concept will be generalized as one relational concept in the high-level module. In our experiments, this parameter is set to an equitable value selected based on some prior knowledge and trial and errors. However, it could be set to a desired value to satisfy the requirements of the application. \autoref{Clustering_fig} illustrates the clustering process when a new demonstration of triangle concept is added to the memory.

\textit{Receiving negative reinforcement signal}: In the case of receiving a negative signal, the robot uses its next most similar untried concept (i.e the one not in $Q_{tried}$), until it receives a positive feedback. If the robot uses all its learned concepts without receiving a positive feedback, then the new observed demonstration will be learned as a novel exemplar of a new concept using \textit{memory rehearsal}.

\subsubsection{No other untried concept exists in the robot's memory}
This situation means that none of the former tried concepts in the robot's memory were proper for the novel demonstration; therefore, a new concept is generated and the new demonstration is consolidated in the robot's memory as an exemplar of that concept through \textit{memory rehearsal}.

\subsubsection{Memory rehearsal}
Memory rehearsal is performed to learn a novel demonstration of a new concept, or to form a novel prototype for an earlier learned concept. Learning new demonstrations faces memory interference which damages previously learned patterns in the memory. This is due to the distributed representation of all patterns in a single network (various patterns share the same synaptic weights in the network). Despite its numerous advantages, memory interference is one of the challenges of employing distributed representation scheme to abstract patterns. To overcome this difficulty, rehearsing and consolidation according to a biological hypothesis is employed \cite{squire1984medial}. 

In the memory rehearsal, previous consolidated prototypes and exemplars in \textit{Mem} are first regenerated using \textit{Mem.TrajectoryNet} as a long-term memory. To do this, the values of PB neurons and initial input neurons in the network's input layer are set to the associated values of the consolidated prototypes or exemplars in \textit{Mem}. Then, the corresponding patterns are regenerated. The regenerated patterns are temporarily stored in a temporal storage called temporal memory. New demonstration is also added to the temporal memory. Next, \textit{Mem.TrajectoryNet} is trained with all the patterns in the temporal memory, starting from the previous network in order to speed up the network's training process. After that, \textit{Mem} is updated based on the new trained \textit{Mem.TrajectoryNet} and the prior associated information of patterns in temporal memory (e.g. \textit{nPrototypes}, \textit{numSamples}, \textit{numSteps}, \textit{initalInfo} and \textit{conceptLables}). Finally, the temporal memory is released.

Like infants in their early years of life, a na{\"\i}ve robot should spend considerable time for learning a sufficient number of patterns through rehearsing and consolidation. In this step, more interactions with teacher are needed to learn concepts during imitation. However, as time passes, the robot has a variety of previously learned concepts in its memory and consequently it responds to the teacher more appropriately with less interactions. But, it is clear that by observing a new concept, the robot should spend time to rehearse and consolidate it. This is similar to the costs and practices that humans experience to learn a new skill.

\subsection{Inference Phase}
In an incremental method, the learning process never stops. However, to assess the performance of ILoCI, an inference phase is designed. In this phase, no further feedbacks are provided by the teacher. When observing a new demonstration, the robot uses its current acquired knowledge during the learning phase to recognize the concept of the new demonstration. $PB_{obs}$ is computed and its value is compared with the values of consolidated $PBs\_rec$ vectors in the memory. The concept of the most similar vector is considered as the concept of the demonstration and a proper action is responded to the teacher.
\section{Results and Discussion}
\label{results_discussion}
To assess the generalization ability of ILoCI in facing large number of concepts and to make it directly comparable with other competing algorithms, its performance is evaluated on a standard benchmark data set, called LASA \cite{khansari2011learning, lemme2015open}. LASA consists of 26 various handwriting motions, collected from pen input using a tablet PC \cite{khansari2011learning, lemme2015open} (supplementary data). All motion shapes constitute 22 distinct relational concepts together in total. It is worthy to note that the shapes are incrementally and gradually demonstrated to the robot to learn their relational concepts while imitating and interacting with the teacher. 

To recognize and generate the observed demonstrations in future, the robot needs to learn the motor data along with the associated observed sensory information. Thereby, the observed teacher's handwriting motion is scaled and fitted in a selected y-z plane in the robot's workspace. The selected workspace, depicted as a supplementary figure, assures the feasibility of executing the action by robot considering its physical limitations and valid workspace \cite{alibeigi2016inverse}. Our test platform is the Aldebaran Robotics$^{\circledR}$ Nao humanoid robot version V3.2 \cite{NaoHumanoidRobot}. After scaling and fitting processes, the joint angles of Nao's right arm will be obtained by applying the built-in inverse kinematics module (IK) on the processed demonstration. To make the results invariant to the possible translational and rotational transformations, the relative displacement values of sensory and motor data are used instead of their absolute values as the inputs to the learning algorithm.

Five-fold cross-validation is used to examine the performance of the proposed algorithm. Each fold consists of different combinations of demonstrations for training and testing. Variant perceptual representations of each shape are randomly divided to five partitions and each of the partitions is used once as training and four times as testing data set. The ideal situation for the robot is to learn the concepts fast while observing only a few numbers of demonstrations and acquiring more comprehensive prototypes. Thus, only 20\% of the demonstrations are used for training and the remaining 80\% are used for testing in each fold. 
In the experiment, \textit{Mem.TrajectoryNet} has 6 input/output nodes, 4 PB neurons, 25 context and 60 hidden neurons. Moreover, $K_{cutoff}$, $Num_{threshold}$ and $Similarity_{threshold}$ are set to 0.5, 3 and 0.1 values, respectively. 

The average correct classification rate over all five folds is $91.346\,\pm\,3.511$ during the inference phase. \autoref{CCR_table} presents the sparse representation of the average normalized confusion and confidence matrices. The full representation of these matrices are available as supplementary data. True positive values in \autoref{CCR_table} show that the robot can correctly recognize demonstrations of each relational concept with high confidence values. Although, some motion shapes in LASA data set have considerable degree of similarity with each other, but the algorithm can discriminate them properly. These similarities also explain the false negative values for some shapes like Line and Saeghe as well as Angle, NShape and Worm. However, the low confidence values for the false negatives indicate that the algorithm is unsure about these results. This ability to properly judge its outcomes brings the metacognition property to the robot. 

\begin{figure}[tpb]	
	\centering
	\includegraphics[width=0.8\linewidth]{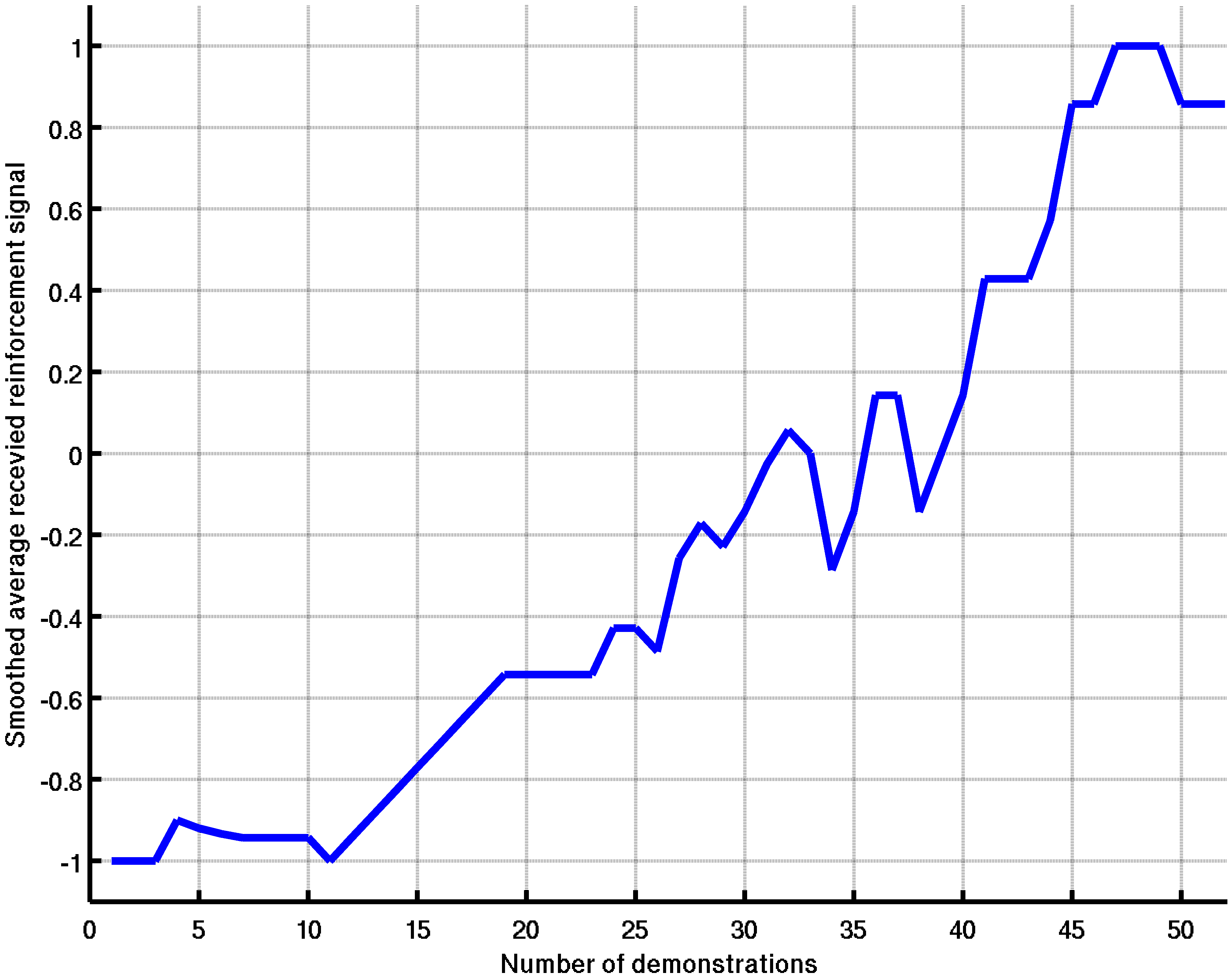}
	\caption{Smoothed average reinforcement signal issued by the teacher during the learning phase in the experimental scenario on LASA data set.}
	\label{Smoothed_RL_fig}
\end{figure}

\begin{table}[tbp]
	\caption{Sparse representation of the average normalized confusion matrix and the corresponding average confidence values on LASA handwriting data set over 5-fold cross-validation. Bold texts indicate true positive values.}
	\label{CCR_table}
	\centering
	\scriptsize 
	\bgroup
	\def\arraystretch{1.2}	
	\resizebox{\linewidth}{!}{%
	\begin{tabular}{|m{0.01\linewidth}|m{0.13\linewidth}|m{0.74\linewidth}|} 
		\cline{3-3}
		\multicolumn{1}{l}{}                                           & \multicolumn{1}{l|}{}                       & \multicolumn{1}{c|}{\textbf{Predicted Concept:(Normalized Confusion, Confidence)}}
			 \\ \hline
		\multicolumn{1}{|c|}{}                                         & \textbf{Angle}                              & \textbf{Angle: (66.67, 1.71)}, Line: (3.33, 0.01), NShape: (13.33, 0.15), Trapezoid: (10, 0.12), Worm: (6.67, 0.07) \\ \hhline{|~|-|-|} 
		\multicolumn{1}{|c|}{}                                         & \cellcolor[HTML]{EFEFEF}\textbf{BendedLine} & \cellcolor[HTML]{EFEFEF}\textbf{BendedLine: (100, 9.12)}   \\ \hhline{|~|-|-|}  
		\multicolumn{1}{|c|}{}                                         & \textbf{CShape}                             & \textbf{CShape: (96.67, 7.08)}, Sshape: (3.33, 0.01) \\ \hhline{|~|-|-|}
		\multicolumn{1}{|c|}{}                                         & \cellcolor[HTML]{EFEFEF}\textbf{GShape}     & \cellcolor[HTML]{EFEFEF}\textbf{GShape: (80, 2.98)}, CShape: (6.67, 0.13), Sshape: (10, 0.04), Worm: (3.33, 0.06)  \\ \hhline{|~|-|-|}
		\multicolumn{1}{|c|}{}                                         & \textbf{JShape}                             & \textbf{JShape: (100, 7.99)}          \\ \hhline{|~|-|-|}
		\multicolumn{1}{|c|}{}                                         & \cellcolor[HTML]{EFEFEF}\textbf{Khamesh}    & \cellcolor[HTML]{EFEFEF}\textbf{Khamesh: (100, 4.31)}                                         \\ \hhline{|~|-|-|}
		\multicolumn{1}{|c|}{}                                         & \textbf{LShape}                             & \textbf{LShape: (96.67, 2.51}, Heee: (3.33, 0.04)    \\ \hhline{|~|-|-|} 
		\multicolumn{1}{|c|}{}                                         & \cellcolor[HTML]{EFEFEF}\textbf{Leaf}       & \cellcolor[HTML]{EFEFEF}\textbf{Leaf: (91.67, 6.52)}, JShape: (1.66, 0.13), Snake: (6.67, 0.12)   \\ \hhline{|~|-|-|} 
		\multicolumn{1}{|c|}{}                                         & \textbf{Line}                               & \textbf{Line: (86.67, 3.13)}, Saeghe: (13.33, 0.21)  \\ \hhline{|~|-|-|} 
		\multicolumn{1}{|c|}{}                                         & \cellcolor[HTML]{EFEFEF}\textbf{NShape}     & \cellcolor[HTML]{EFEFEF}\textbf{NShape: (60, 1.65)}, Angle: (6.67, 0.07), Worm: (33.33, 0.68)  \\ \hhline{|~|-|-|} 
		\multicolumn{1}{|c|}{}                                         & \textbf{Pshape}                             & \textbf{PShape: (96.67, 2.87)}, Trapezoid: (3.33, 0.02)  \\ \hhline{|~|-|-|} 
		\multicolumn{1}{|c|}{}                                         & \cellcolor[HTML]{EFEFEF}\textbf{RShape}     & \cellcolor[HTML]{EFEFEF}\textbf{RShape: (100, 4.05)}     \\ \hhline{|~|-|-|} 
		\multicolumn{1}{|c|}{}                                         & \textbf{Saeghe}                             & \textbf{Saeghe: (80, 3.32)}, Line: (20, 0.53)  \\ \hhline{|~|-|-|} 
		\multicolumn{1}{|c|}{}                                         & \cellcolor[HTML]{EFEFEF}\textbf{Sine}       & \cellcolor[HTML]{EFEFEF}\textbf{Sine: (100, 3.69)}    \\ \hhline{|~|-|-|} 
		\multicolumn{1}{|c|}{}                                         & \textbf{Snake}                              & \textbf{Snake: (100, 4.34)}  \\ \hhline{|~|-|-|} 
		\multicolumn{1}{|c|}{}                                         & \cellcolor[HTML]{EFEFEF}\textbf{Spoon}      & \cellcolor[HTML]{EFEFEF}\textbf{Spoon: (95, 3.54)}, Heee: (5, 0.04)   \\ \hhline{|~|-|-|} 
		\multicolumn{1}{|c|}{}                                         & \textbf{Sshape}                             & \textbf{Sshape: (90, 2.96)}, GShape: (10, 0.03)  \\ \hhline{|~|-|-|} 
		\multicolumn{1}{|c|}{}                                         & \cellcolor[HTML]{EFEFEF}\textbf{Trapezoid}  & \cellcolor[HTML]{EFEFEF}\textbf{Trapezoid: (100, 3.88)}   \\ \hhline{|~|-|-|} 
		\multicolumn{1}{|c|}{}                                         & \textbf{WShape}                             & \textbf{WShape: (95, 3.20)}, Khamesh: (5, 0.03) \\ \hhline{|~|-|-|} 
		\multicolumn{1}{|c|}{}                                         & \cellcolor[HTML]{EFEFEF}\textbf{Worm}       & \cellcolor[HTML]{EFEFEF}\textbf{Worm: (86.67, 2.61)}, NShape: (13.33, 0.49)  \\ \hhline{|~|-|-|}
		\multicolumn{1}{|c|}{}                                         & \textbf{ZShape}                             & \textbf{ZShape: (100, 3.56)}  \\ \hhline{|~|-|-|}
		\parbox[t]{2mm}{\multirow{-22}{*}{\rotatebox[origin=c]{90}{\textbf{Actual Concept}}}} & \cellcolor[HTML]{EFEFEF}\textbf{Heee}       & \cellcolor[HTML]{EFEFEF}\textbf{Heee: (65, 2.02)}, LShape: (10, 0.10), Spoon: (3.33, 0.15), ZShape: (21.67, 0.21)   \\ \hline
	\end{tabular}
	}
}
\end{table}
Moreover, to assess the learning speed and the interaction quality of the proposed algorithm, the reinforcement signals given by the teacher during the learning phase is investigated. \autoref{Smoothed_RL_fig} shows the average reinforcement signals (over five folds) given by the teacher. Because of the discrete nature of the reinforcement signals (+1 for reward and -1 for punishment), the results in \autoref{Smoothed_RL_fig} has been smoothed with a backward moving average with window length of seven to reflect the expected behavior clearly. Results show that robot is capable of learning the relational concepts of the observed demonstrations very fast especially at the initial stages of learning. According to \autoref{Smoothed_RL_fig}, in 85\% (in average) of the experiments, the robot has correctly recognized the relational concepts in the first interaction after merely learning 45 demonstrations (25\% of the data). 
Two specific reasons can be cited for this notable property. First, when a new demonstration with a novel relational concept is observed, it will be consolidated and probably updated later in the memory as a representative of the perceived relational concept. Consequently, the robot has at least one representation for each relational concept in the memory due to the functional abstraction. Therefore, it can recognize new demonstrations quickly using prototypes in its memory without relearning them from scratch. Second, all consolidated exemplars and prototypes are stored in one memory (distributed representation) through memory rehearsal which brings about the utilization of their common structural relations in order to expedite and enhance the learning process.

\begin{figure}[tpb]	
	\centering
	\includegraphics[width=\linewidth]{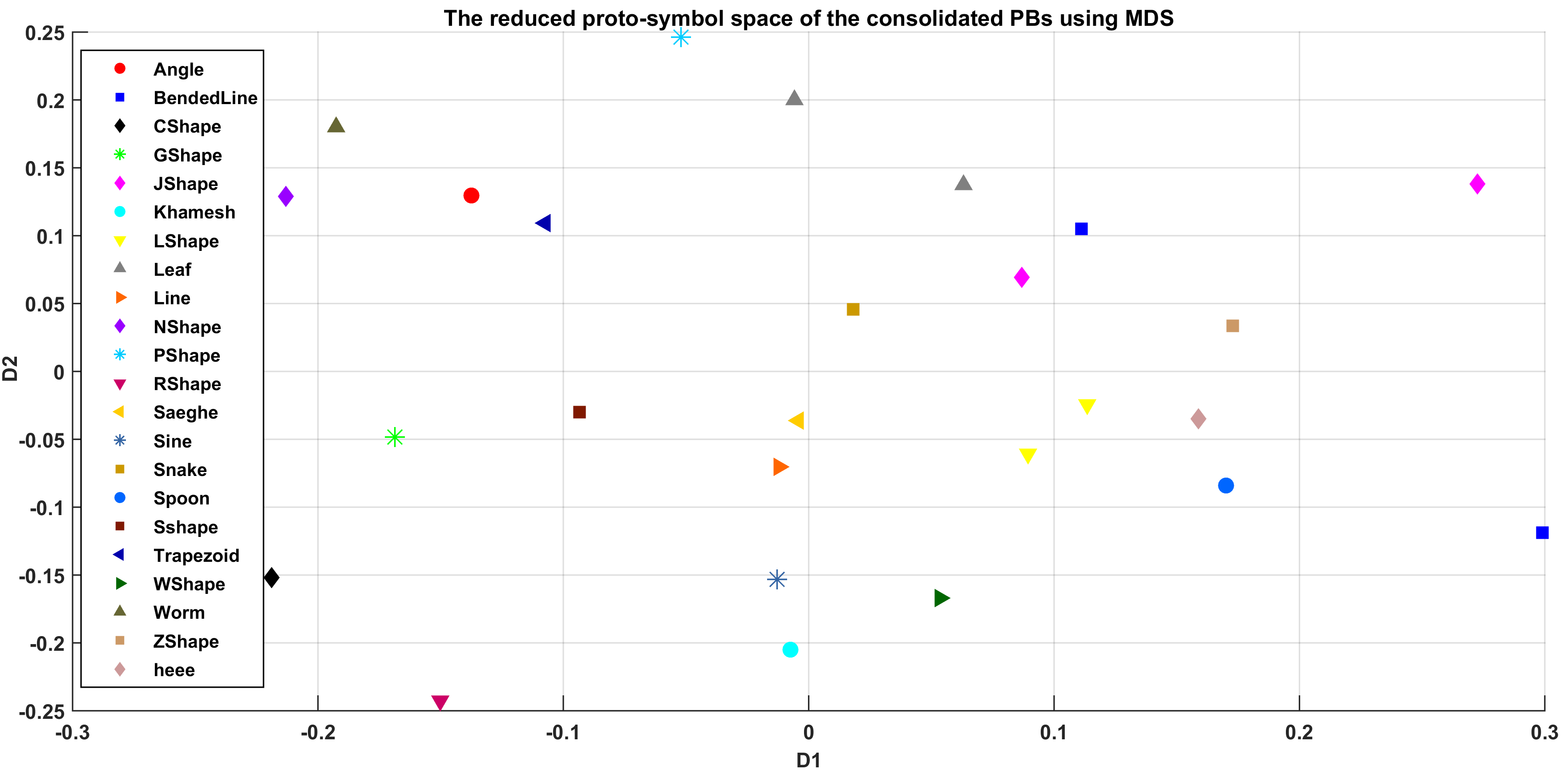}
	\caption{2D visualization of the PB space of the consolidated perceptual prototypes in the fifth fold.}
	\label{PBs_DMS_2D_fig}
\end{figure}

Furthermore, ILoCI unites all different perceptual prototypes of each relational concept in the high-level module based on the teachers' feedbacks. \autoref{PBs_DMS_2D_fig} shows the symbol space (PB space) of the acquired perceptual prototypes in the fifth fold using non-metric multidimensional scaling (MDS). The figure shows the 2D visualization of the acquired 4D PB vectors. In \autoref{PBs_DMS_2D_fig}, all PB vectors associating with different perceptual prototypes of one relational concept are represented with same markers, which shows their unity as one relational concept (e.g. both acquired perceptual prototypes of BendedLine are shown with blue square markers). The results also show that ILoCI almost finds the same number of perceptual prototypes for each relational concept as the number of their real perceptual variants. However, two different perceptual prototypes are acquired here for LShape which has only one distinct perceptual representation since the teachers can draw shapes freely. So, the observed demonstrations might vary and consequently two different perceptual prototypes are formed for LShape. However, it is notable that all variant perceptual prototypes of each relational concept are unified in the high-level module through the functional abstraction. 

\begin{figure}[tpb]	
	\centering
	\includegraphics[width=0.9\linewidth]{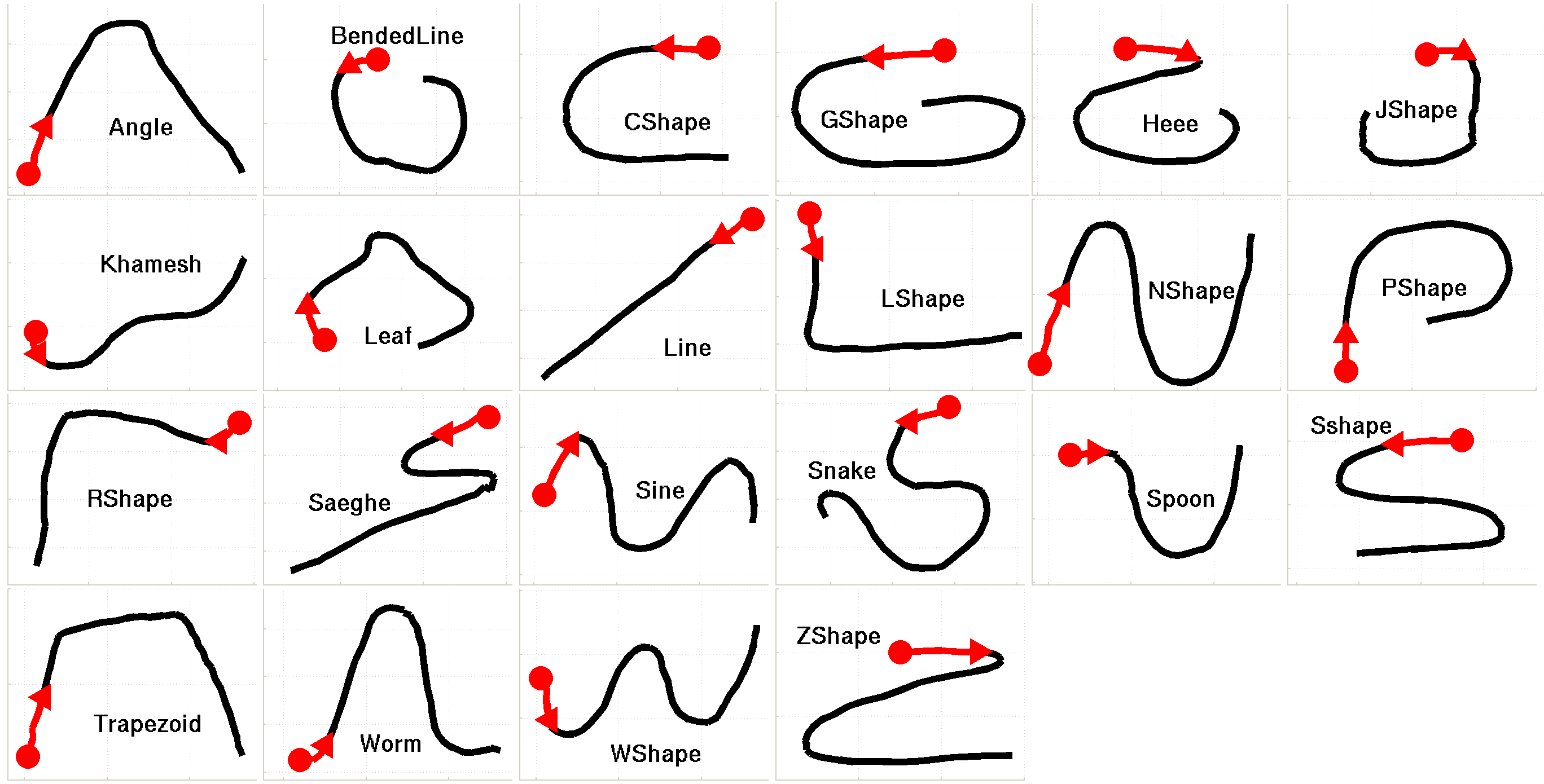}
	\caption{Regenerated samples for each relational concept in the fifth fold. The red circles show the starting points.}
	\label{GeneratedShapes_fig}
\end{figure}

In addition, the proposed algorithm generates smooth and comprehensive prototypes for each relational concept, despite the discrepancies in the observed demonstrations, without any smoothing post-processing. \autoref{GeneratedShapes_fig} shows one regenerated example for each acquired relational concept by the robot. The smoothness of the generated prototypes supports the generalization ability of the algorithm. The supplementary videos show the execution of some presented motions by Nao humanoid robot. 
\section{CONCLUSIONS}
This paper introduced an incremental and gradual model for learning concepts by imitation as one of the manifestations of true imitation learning. The presented algorithm autonomously and incrementally learns concepts from observed multimodal spatio-temporal demonstrations, based on both their perceptual and functional properties during imitation. It abstracts demonstrations both at the trajectory and the symbolic levels, which is a significant challenge in integrating the symbolic AI and the continuous control of robots \cite{billard2008handbook}. In this method, all perceptual concepts are incrementally learned in a single recurrent neural network through the proposed memory rehearsal. Functional similarities between concepts are also acquired through a limited number of interactions with the teacher. Incremental learning of acquired concepts together through memory rehearsal enables robot to utilize the common structural relations among demonstrations. Consequently the learning process is expedited especially at the initial stages and the generalization ability of the algorithm is also increased. 

The performance of the proposed method was assessed using standard LASA benchmark data set \cite{khansari2011learning, lemme2015open}. Results show that due to abstraction and generalization in both perceptual and functional spaces, robot acquires comprehensive prototypes and therefore it can truly recognize concepts of observed demonstrations during the imitation. The mentioned properties make the proposed method a good choice for real-world applications in which robots should comprehend intentions of their partners while interacting with them.

\section{Supplementary Material}
The supplementary material is available at \url{https://goo.gl/ojowSx}.
\addtolength{\textheight}{-12cm}   



%

%
%
%
%
%
%
%
{
\small
\singlespacing
\setlength{\itemsep}{-2ex}
\bibliographystyle{ieeetr}
\bibliography{ms}
}

\end{document}